\documentclass[11pt]{article}

\usepackage[final]{acl}

\usepackage{times}
\usepackage{latexsym}

\usepackage[T1]{fontenc}

\usepackage[utf8]{inputenc}

\usepackage{microtype}

\usepackage{inconsolata}

\usepackage{graphicx}

\usepackage{enumitem}
\usepackage{graphicx}
\usepackage[utf8]{inputenc} 
\usepackage[T1]{fontenc}    
\usepackage{hyperref}       
\usepackage{url}            
\usepackage{booktabs}       
\usepackage{amsfonts}       
\usepackage{nicefrac}       
\usepackage{microtype}      
\usepackage{xcolor}         
\usepackage[table]{xcolor}
\usepackage{array}
\usepackage{amssymb}
\usepackage{enumitem}
\usepackage{graphicx}
\usepackage{subcaption}
\usepackage{threeparttable}
\usepackage{pifont}
\usepackage{amsmath}
\usepackage{multirow}
\newcommand{\xmark}{\ding{55}}
\usepackage{listings}
\usepackage{makecell}
\usepackage{hyperref}
\usepackage{url}
\usepackage{fontawesome5}

\definecolor{red}{RGB}{238, 68, 51}
\definecolor{blue}{RGB}{70, 177, 225}
\definecolor{yellow}{RGB}{255, 192, 0}
\definecolor{purple}{RGB}{216, 110, 204}
\definecolor{brown}{RGB}{127, 36, 28}
\definecolor{green}{RGB}{71, 172, 20}
\definecolor{orange}{RGB}{194,153,107}

\newcolumntype{C}[1]{>{\centering\let\newline\\\arraybackslash\hspace{0pt}}m{#1}}

\title{A3: Android Agent Arena for Mobile GUI Agents with Essential-State Procedural Evaluation}

\author{
 \textbf{Yuxiang Chai\textsuperscript{1,4}\textsuperscript{*}},
 \textbf{Shunye Tang\textsuperscript{2,4}\textsuperscript{*}},
 \textbf{Han Xiao\textsuperscript{1,4}},
 \textbf{Weifeng Lin\textsuperscript{1,4}},
 \textbf{Hanhao Li\textsuperscript{3}},
 \\
 \textbf{Jiayu Zhang\textsuperscript{3}},
 \textbf{Liang Liu\textsuperscript{4}},
 \textbf{Pengxiang Zhao\textsuperscript{4}},
 \textbf{Guangyi Liu\textsuperscript{4}},
 \textbf{Guozhi Wang\textsuperscript{4}},
 \\
 \textbf{Shuai Ren\textsuperscript{4}},
 \textbf{Rongduo Han\textsuperscript{2}},
 \textbf{Haining Zhang\textsuperscript{2}},
 \textbf{Siyuan Huang\textsuperscript{5}},
 \textbf{Hongsheng Li\textsuperscript{1}\textsuperscript{\textdagger}},
\\
 \small
 \textsuperscript{*}Equal contribution,
 \textsuperscript{1}MMLab @ CUHK,
 \textsuperscript{2}Nankai University,
 \textsuperscript{3}EE @ CUHK,
 \textsuperscript{4}vivo AI Lab,
 \textsuperscript{5}SJTU
 \\
 \small
 \faGithub \quad \url{https://github.com/YuxiangChai/AITK}
}

\begin{document}
\maketitle

\begin{abstract}
The advancement of Large Language Models (LLMs) and Multimodal Large Language Models (MLLMs) has catalyzed the development of mobile graphic user interface (GUI) AI agents, which is designed to autonomously perform tasks on mobile devices. However, a significant gap persists in mobile GUI agent evaluation, where existing benchmarks predominantly rely on either static frame assessments such as AndroidControl or offline static apps such as AndroidWorld and thus fail to capture agent performance in dynamic, real-world online mobile apps. To address this gap, we present Android Agent Arena (A3), a novel "essential-state" based procedural evaluation system for mobile GUI agents. A3 introduces a benchmark of 100 tasks derived from 20 widely-used, dynamic online apps across 20 categories from the Google Play Store, ensuring evaluation comprehension. A3 also presents a novel "essential-state" based procedural evaluation method that leverages MLLMs as reward models to progressively verify task completion and process achievement. This evaluation approach address the limitations of traditional function based evaluation methods on online dynamic apps. Furthermore, A3 includes a toolkit to streamline Android device interaction, reset online environment and apps and facilitate data collection from both human and agent demonstrations. The complete A3 system, including the benchmark and tools, will be publicly released to provide a robust foundation for future research and development in mobile GUI agents.
\end{abstract}

\section{Introduction}

The rapid evolution of Large Language Models (LLMs) and Multimodal Large Language Models (MLLMs) has acted as a primary catalyst for innovation in autonomous AI agents. Within this domain, Graphical User Interface (GUI) agents represent a critical area of research. Unlike systems that interact with applications via APIs or textual representations (e.g., HTML and XML), GUI agents operate through a fundamentally visual modality, executing tasks by directly processing screen pixel information. This paper focuses on mobile GUI agents, which are specialized for the interaction paradigms of the mobile ecosystem. These agents are becoming increasingly prevalent due to their potential to autonomously execute user commands, thereby streamlining workflows and minimizing human intervention.

While research of mobile GUI agents has advanced in modeling, evaluation remains a persistent challenge. The inherently sequential nature of tasks makes ascertaining task success difficult. Pioneering benchmarks~\citep{chai-etal-2025-amex,rawles2023androidinthewild,li2024effects} address this via static frame-step evaluation, assessing an agent's ability to predict the next action from a single static screenshot. Although this successfully measures single-step accuracy, it fails to simulate real-world fidelity, where it cannot account for dynamic state changes or the cascading effects of early errors that often lead to task failure. Furthermore, static evaluation penalizes valid alternative trajectories. For instance, launching an app via the library versus using ADB commands are both correct but static evaluation would treat one of them as failure.

\begin{figure*}
    \centering
    \includegraphics[width=0.8\linewidth]{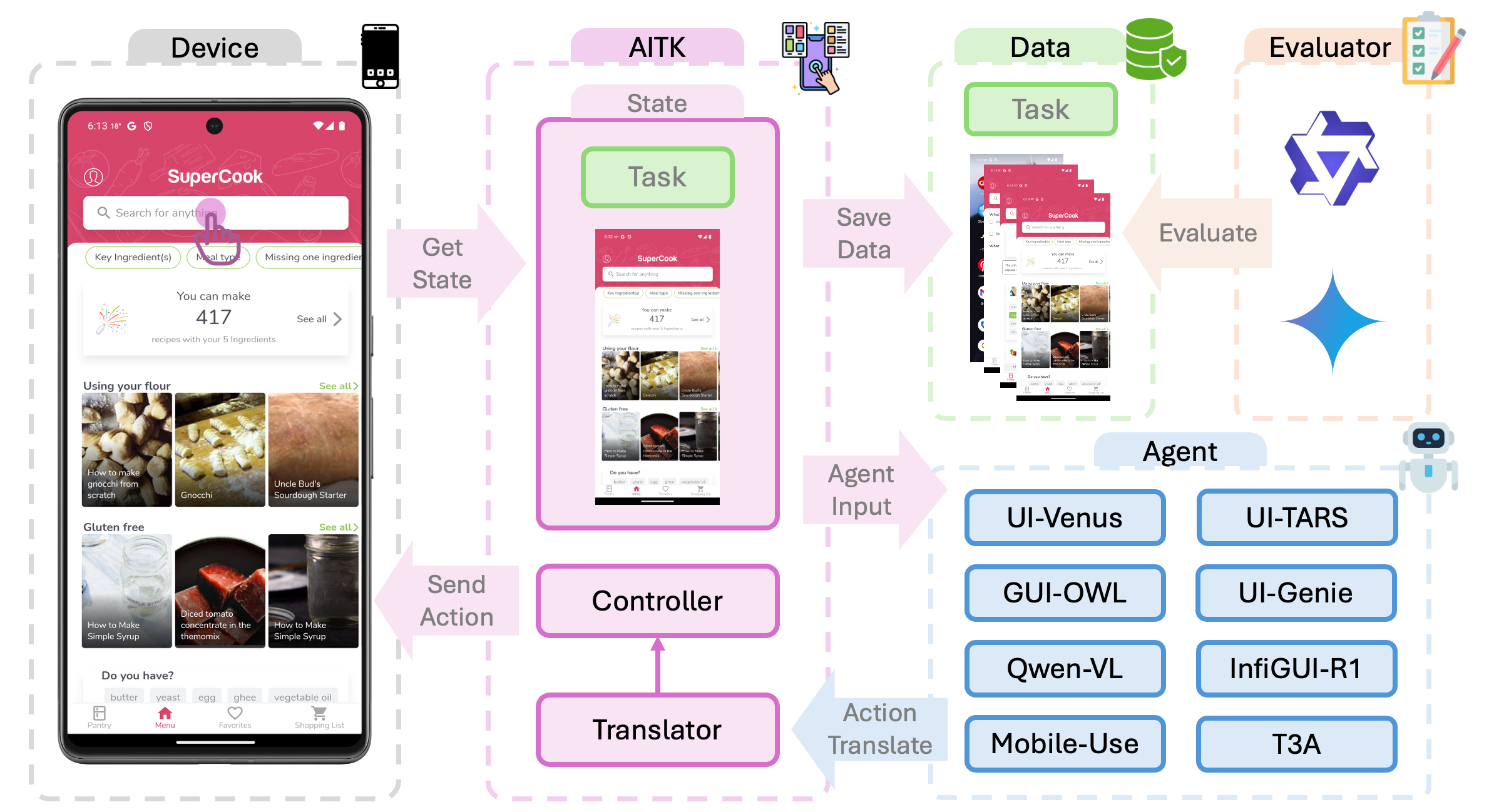}
    \caption{A3 consists of AITK (Android Interaction Toolkit), agents and evaluators. \textcolor{purple}{AITK} has a translator to convert \textcolor{blue}{agent} actions to a unified action space and a controller to interact with the \textcolor{gray}{device}. AITK also gets the state of device at each step and save the \textcolor{green}{data}. After the task execution, \textcolor{orange}{evaluator} output the evaluation results.}
    \label{fig:pipeline_suite}
\end{figure*}

To overcome the limitations of static assessment, the field has shifted toward dynamic evaluation systems using interactive devices. A prominent example, AndroidWorld~\citep{rawlesandroidworld}, evaluates agents by instrumenting the source code of open-source apps to verify internal states in an interactive virtual device. However, this reliance on open-source software creates a significant gap in ecological validity: widely used closed-source applications in categories such as shopping, travel, news, ticketing, etc. are excluded. Other benchmarks~\citep{xu-etal-2025-androidlab, lee2025benchmarkingmobiledevicecontrol} share the same shortcomings. Consequently, agents are rarely tested on the apps users interact with most. Moreover, some dynamic frameworks~\citep{chen2024spa, xing2024androidarena} suffer from environmental instability, difficult state resets, and high reliance on manual human evaluation, leading to scalability issues and potential inaccuracies.

To address these shortcomings, we propose the Android Agent Arena (A3). Our evaluation system introduces a comprehensive benchmark comprising 100 daily-life tasks derived from 20 popular applications spanning 20 categories in the Google Play Store's top charts. Crucially, A3 incorporates dynamic and online apps previously deemed untestable due to the constraints of function-based evaluation, such as news, navigation, travel, email, shopping, etc. Because the internal states of these constantly updating proprietary apps cannot be instrumented, we introduce a novel \textbf{essential-state based procedural evaluation method}. This approach leverages MLLMs, either large commercial models or samller fine-tuned open-source alternatives, as reward models to autonomously and progressively determine task success and essential state achievements. This methodology not only verifies final objectives but also evaluates intermediate progress even when the total task is not completed, significantly reducing manual labor while maintaining high evaluation reliability on dynamic online apps and tasks. A3 pipeline is demonstrated in Figure~\ref{fig:pipeline_suite}.

Our contributions are summarized as follows: 
\begin{itemize}[leftmargin=10pt]
\vspace{-0.6em}
\item We introduce the Android Agent Arena (A3), a benchmark featuring 100 common tasks derived from 20 popular, dynamic online apps. This enables the robust evaluation of agent performance in complex, real-world scenarios that were previously difficult to assess.
\vspace{-0.6em}
\item We propose a novel essential-state procedural evaluation methodology that utilizes MLLMs to progressively verify task success. We demonstrate that both large commercial models and our fine-tuned lightweight alternatives (A3RM) can serve as effective reward models to assess intermediate progress and final objectives.
\vspace{-0.6em}
\item We release the complete pipeline and tools to accelerate research in the field. This includes modules for streamlined agent execution, trajectory data collection (for both agents and humans), and a versatile evaluator module designed for easy customization and community adoption.
\end{itemize}

\section{Related Work}

\begin{table*}[ht]
    \centering
    \resizebox{0.98\linewidth}{!}{
    \begin{tabular}{l C{2cm} C{2cm} C{2.5cm} C{2.5cm} C{2.5cm} C{2cm}}
        \toprule
        \textbf{Name} & \textbf{Eval Mode} & \textbf{\# Tasks}  & \textbf{\# General Apps} & \textbf{Operation} & \textbf{Inf. Query} & \textbf{Online} \\
        \midrule
        \textsc{AitW} & static & -  & - & \checkmark & \xmark & \xmark \\
        AndroidControl & static  & - & - & \checkmark & \xmark & \xmark \\
        AMEX & static & -  & - & \checkmark & \xmark & \xmark \\
        GUI-Odyssey & static & - & - & \checkmark & \xmark & \xmark \\
        \midrule

        AndroidArena & dynamic & 221  & 4 & \checkmark & \xmark & \xmark\\
        Mobile-Env & dynamic & 74  & 5 & \checkmark & \xmark & \xmark \\
        AndroidWorld & dynamic & 116  & 15 & \checkmark & \checkmark & \xmark\\
        B-Moca & dynamic & 131 & 4  & \checkmark & \xmark & \xmark \\
        AndroidLab & dynamic & 138  & 5 & \checkmark & \checkmark & \xmark \\
        SPA-bench & dynamic & 170 & 20 & \checkmark & \xmark & \checkmark \\
        \midrule

        A3 (Our) & dynamic & 100  & 20 & \checkmark & \checkmark & \checkmark\\
        \bottomrule
    \end{tabular}
    }
    \caption{GUI related datasets and benchmarks in English. The top four rows are GUI agent related datasets, which provide static frame evaluation. The middle six rows are dynamic evaluation systems, which provide different tasks from different apps in different settings. AndroidWorld provides 15 generals apps from non-mainstream open-source F-Droid. SPA-bench provides another 20 apps and 170 tasks in Chinese.}
    \label{tab:compare-benchmark}
\end{table*}

\subsection{GUI Agents}
Recent advancements increasingly leverage the world knowledge and reasoning capabilities of modern MLLMs for GUI control tasks, aiming to build more general and autonomous interactive agents \citep{liu2025llmpowered, wang2025guiagentsfoundationmodels, hu-etal-2025-os}. Existing approaches can be broadly grouped into two categories. The first line finetunes a single general-purpose base MLLM~\citep{bai2025qwen25vltechnicalreport} as the GUI agent, relying on their unified visual-textual understanding to directly plan and execute actions in diverse interfaces~\citep{qin2025ui-tars,gu2025uivenustechnicalreportbuilding,liu2025infiguir1advancingmultimodalgui,xiao2025uigenie}. The second line comprises framework style systems like Mobile-Use~\citep{li2025mobileuse} that integrate GUI-specific perception or UI-tree modules, multi-agent planning or shortcut guidance to improve robustness and task efficiency.

\subsection{GUI Agent Benchmarks}

Early evaluations of GUI agents primarily relied on static benchmarks where an agent predicts the next action from a single screenshot. Prior works such as AITW \citep{rawles2023androidinthewild}, AMEX \citep{chai-etal-2025-amex}, and AndroidControl \citep{li2024effects} focused on single-step action accuracy via element or coordinate matching. While valuable for step-level assessment, this static evaluation fails to capture an agent's ability to perform sequential actions in dynamic, interactive environments. This limitation motivated the development of dynamic evaluation frameworks that are the focus of our work.

However, existing dynamic benchmarks exhibit significant constraints in their task and application design. Several systems are restricted by task simplicity and diversity, such as Mobile-Env \citep{zhang2023mobileenv} and B-Moca \citep{lee2025benchmarkingmobiledevicecontrol}. Others are constrained by their app selection: AndroidArena \citep{xing2024androidarena}, for instance, focuses on system apps, failing to evaluate generalization to the third-party app ecosystem. Similarly, prominent benchmarks like AndroidWorld \citep{rawlesandroidworld} and AndroidLab \citep{xu-etal-2025-androidlab} are restricted to open-source and offline apps. These benchmarks exclude common real-world stochastic events such as pop-up ads and dynamic content updates, that are critical for testing agent robustness. They also omit online dynamic apps such as shopping, travel, and news, which are common in real-life scenarios. While SPA-bench \citep{chen2024spa} includes online apps, it suffers from instability and unreliable system resets. A more foundational issue across many of these benchmarks is the reliance on simplistic evaluation methodologies. Information query success is often determined by matching predefined answers \citep{xu-etal-2025-androidlab}, and operational success by exact state matching \citep{rawlesandroidworld}. These rigid, programmatic checks fail to capture the nuances of task completion in dynamic environments. This highlights a critical need for a benchmark with ecologically valid tasks coupled with a more flexible, semantically-aware evaluation framework. The overall statistics of the existing benchmarks are listed in Table~\ref{tab:compare-benchmark}.

\section{Android Agent Arena (A3)}

\subsection{Apps \& Tasks}

\begin{figure}
    \centering
    \includegraphics[width=0.98\linewidth]{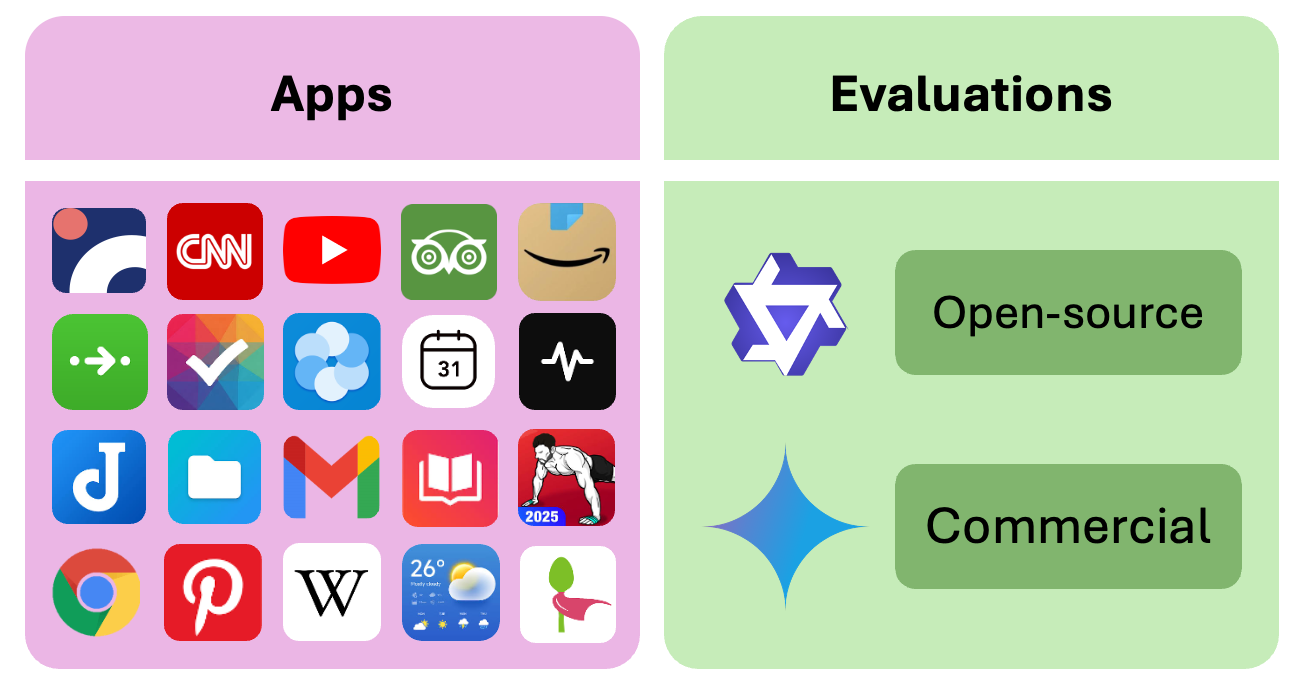}
    \caption{Overview of A3 apps and evaluations. The benchmark features tasks from 20 popular mobile applications and the framework supports essential-state based evaluation by two MLLM-based methods, utilizing either a commercial or an open-source finetuned model.}
    \label{fig:tasks}
\end{figure}

Existing benchmarks exhibit significant limitations regarding ecological validity and reproducibility. Many rely on offline, open-source applications with restricted functionality~\citep{rawlesandroidworld,xu-etal-2025-androidlab}. While these environments enable stable testing, they prioritize low-complexity system utilities, such as starting a timer or starting a voice recording which are often trivially solvable by API-based assistants (e.g., Siri), or drawing and moving boxes on a static HTML canvas which users would never ask agents to do in the real-life. In contrast, we consider that robust mobile GUI agents should address user intents in common apps with in-the-wild tasks, such as playing a tutorial video when cooking, searching for the cheapest flight or finding the fast route when navigating. While ~\citet{chen2024spa} attempts to incorporate online apps, it fails to provide reproducible test environments or automated device-reset mechanisms. This lack of standardization necessitates extensive human intervention for initialization and cleanup, causing large-scale, reproducible testing prohibitive.

To address these challenges, we adopted a systematic top-down curation strategy. First, we identified 20 distinct categories from the Google Play Store top charts to ensure broad domain coverage. Within each category, we evaluated multiple candidate applications using a rigorous filtering protocol to select one representative app. Our final selection of 20 applications, averaging 115 million downloads each (see Figure~\ref{fig:tasks} and Appendix~\ref{app:app_list}), was guided by a critical technical criterion: feasibility for automated state management, specifically requiring minimal forced logins and supporting reliable environment resets. This curation resolves the fundamental trade-off between ecological validity and experimental reproducibility. A3 establishes a unique framework that supports dynamic, online apps while maintaining reliable, programmatic resets to a consistent initial distribution, enabling the robust evaluation of agents within a stable experimental setting.

Building on this foundation, we designed 100 representative tasks aligned with the core functionalities of the selected applications (e.g., restricting Amazon tasks to shopping and CNN tasks to news retrieval). To provide a structured evaluation, tasks are classified along two primary axes. Based on their objective, they are designated as either Operation (requiring a sequence of state-changing actions) or Information Query (which additionally requires the agent to extract information to answer a question). The difficulty of each task is quantitatively defined by the number of steps ($N$) required by a human expert: Easy ($N <7$), Medium ($7\leq N\leq 11$), and Hard ($N>11$).

\subsection{Essential-State Evaluation}

Existing evaluation methodologies for mobile agents often rely on metrics that lack sufficient granularity and flexibility for in-depth analysis. The widely-used AndroidWorld benchmark~\citep{rawlesandroidworld}, for instance, primarily employs a binary Success Rate (SR), calculated as the ratio of successful tasks to the total ($N_{success} / N_{total}$). While straightforward, this coarse-grained metric provides no insight into partial progress or specific failure modes, and it cannot differentiate the capabilities of agents that achieve the similar final score. To address this, AndroidLab~\citep{xu-etal-2025-androidlab} introduced an additional, more granular sub-goal success rate. While this approach offers a more detailed view by tracking intermediate steps, its implementation has significant limitations. The sub-goals are pre-defined in a rigid JSON format (e.g., \textit{Phone: 12345678} ) for the evaluation function and are only applicable to certain operation tasks.

To address the need for a granular yet flexible metric, we introduce the core of A3's framework: the essential-state evaluation method. We define an essential state as a critical, observable semantic milestone that must be achieved for a task to be considered successful. Instead of assessing a binary final outcome, our method decomposes the execution trajectory into a sequence of these verifiable states. This approach differs fundamentally from the rigid sub-goals employed in systems like AndroidLab~\citep{xu-etal-2025-androidlab}. While AndroidLab sub-goals are typically bound to specific internal element key-value pairs (e.g., exact view IDs or string matches), essential states are defined by higher-level semantic outcomes. This abstraction makes our metric and evaluation resilient to UI variations and diverse execution paths, which is crucial for dynamic apps and open-ended information query tasks. For example, consider the task: \textit{"Search 'marvel comics' in Pinterest. Who is the author of the first pin?"} We define three essential states: (1) The query 'marvel comics' is searched; (2) The first pin is selected; and (3) The author of the pin is identified. Crucially, because these states are defined by what is achieved rather than how the underlying code renders it, they remain invariant even as app content dynamically refreshes or evolves. A detailed list of tasks and their corresponding essential states is provided in Appendix~\ref{app:tasks}.

Defining such semantic milestones in human-centric tasks inherently involves qualitative judgment. To ensure rigor and mitigate annotator bias, we established a strict three-stage, human-in-the-loop protocol for defining these states:
\begin{itemize}[leftmargin=10pt] 
\vspace{-0.3em}
\item \textbf{Trajectory Diversity:} Two human operators independently complete the same task using distinct strategies to generate diverse successful trajectories.
\vspace{-0.6em}
\item \textbf{Collaborative Definition:} Based on the collected trajectories, the operators collaboratively propose a set of essential states. These states must meet three criteria: (1) \textit{Visual Verifiability} (identifiable from consecutive screenshots and actions); (2) \textit{Criticality} (representing a "must-be-done" step); and (3) \textit{Sufficiency} (the set must cover the entire task logic).
\vspace{-0.6em}
\item \textbf{Independent Audit:} A human evaluator audits the proposed states to verify they are logical, comprehensive, and achievable across different valid interaction methods.
\end{itemize}

\vspace{-0.2em}
We acknowledge that defining essential states incurs an initial manual design cost. However, this represents a strictly one-time investment. As established, essential states capture high-level task logic rather than transient UI elements or specific pixel coordinates. Consequently, these states remain invariant as long as the core task objective is unchanged, effectively decoupling the evaluation logic from the dynamic content refreshes inherent to online applications. This stability ensures that the benchmark remains valid over time without frequent re-annotation, offering a long-term evaluation reliability.

For essential-state evaluation metric, we define $ESAR = N_{AES} / N_{TES}$, where $ESAR$ is the \textbf{E}ssential-\textbf{S}tate \textbf{A}chieved \textbf{R}ate, $N_{AES}$ is the number of achieved essential-states and $N_{TES}$ is the total number of essential-states. Overall task success is then determined by the successful completion of all its associated essential states. We formalize this as:
\begin{equation*}
    A_i = \begin{cases}
        1\text{, \;\;if}\;\;eval(ES_j) = 1 \;\; \\ \;\;\;\;\;\;\;\;\;\;\text{for}\;\;\forall j \in [0,...,N_{ES}] \\
        0\text{, \;\;otherwise}
    \end{cases}
\end{equation*}
where $A_i$ is the success status of task $i$, $ES_j$ is the $j$-th essential-state for the task, $N_{ES}$ is the total number of the essential-states for task $i$ and $eval(\cdot)$ is the method that verifies the achievement of an essential-state. Thus, the complex problem of end-to-end task evaluation is simplified into a series of more manageable essential-state assessments.

\begin{figure}
    \centering
    \includegraphics[width=1\linewidth]{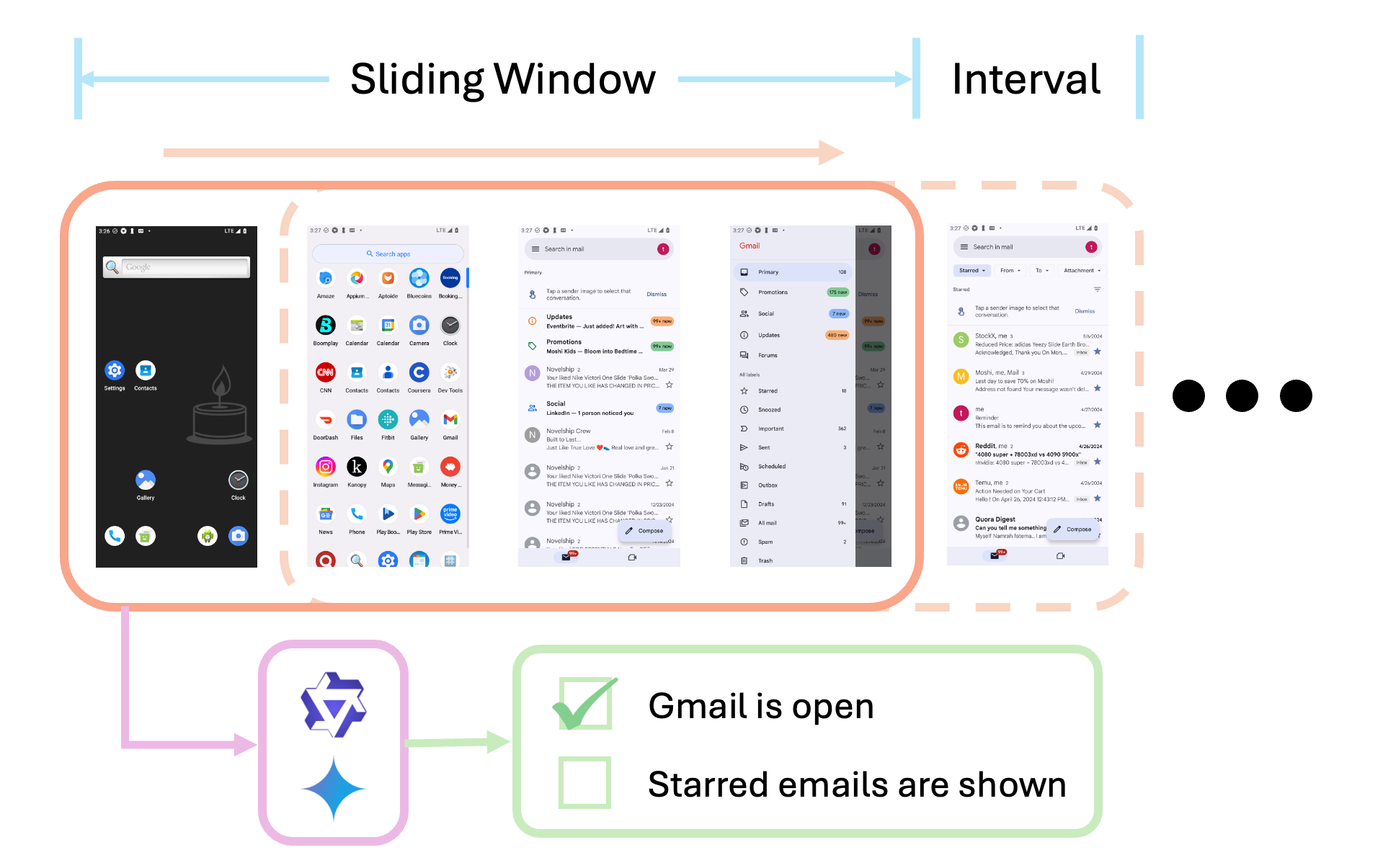}
    \caption{Illustration of the sliding window mechanism. A window of a predefined size and interval traverses the trajectory. At each position, the screen frames within the window are passed to a judge. The judge then assesses whether any of the task's essential states have been achieved within that segment. This process repeats until the entire trajectory has been evaluated.}
    \label{fig:sliding_window}
\end{figure}

To implement the essential-state evaluation, we process the agent's full interaction trajectory using a sliding window mechanism. A window of a predefined size traverses the sequence of screen frames from start to finish, advancing at a specified interval. At each position, the contents of the window are submitted to our MLLM-based evaluator, which judges whether any of the essential-states are satisfied within that specific segment. This evaluation process is illustrated in Figure~\ref{fig:sliding_window}.

\subsection{MLLM-Based Judgment}

A critical challenge in our framework is how to verify the achievement of each essential state. Traditional function-based methods, which rely on parsing UI structures like the XML or View Hierarchy, are not suitable for this task of dynamic, constantly updating apps. These methods  often struggle for two primary reasons: (1) many third-party proprietary apps feature mis-aligned or inconsistent UI trees that are difficult to parse reliably, and (2) they suffer from a semantic gap, as many essential states are defined by high-level concepts (e.g., \textit{"the first product is selected"}) that cannot be verified by simply checking for a specific widget ID or text string in an constantly updating environment.

To overcome these limitations, we leverage the sophisticated visual and language understanding capabilities of modern MLLMs. Inspired by works on using MLLMs as evaluators \citep{chen2024mllmasajudgeassessingmultimodalllmasajudge}, we employ an MLLM to act as the core judgment mechanism for A3. Our approach, however, refines how the MLLM is applied. Prior work like SPA-bench \citep{chen2024spa} pioneered the use of an MLLM to directly evaluate an entire, multi-step task trajectory based on the screenshots combination. This naive end-to-end judgment is a complex reasoning task that resulted in an accuracy of approximately only 80\% in trajectory evaluations, which is unreliable. In contrast, our essential-state evaluation method is perfectly designed to address this challenge by decomposing the overall task into a series of simpler, more discrete verification steps of essential-states. This simplifies the reasoning required of the MLLM, as it makes a focused judgment on a single, much simpler milestone rather than a long sequence, thereby aiming for higher and more reliable evaluation accuracy.

\paragraph{Commercial MLLMs} The advanced capabilities of commercial MLLMs, such as Gemini-2.5-pro, make them highly effective for the role of an automated evaluator. Their robust performance in visual comprehension and instruction following provides a reliable evaluation for judging whether an essential-state has been achieved within a given set of screen frames. Though commercial MLLMs performs good as an evaluator, a significant practical challenge associated with these commercial models is the monetary cost of their API services, as each evaluation call incurs a fee. This creates a critical trade-off between evaluation granularity and cost. Therefore, to ensure our framework is both accurate and economically viable, we find an optimal sliding window parameters, specifically the window size and interval size, to minimize the total number of API calls without compromising evaluation fidelity (experiments and results detailed in Appendix~\ref{app:sliding_window}).

\paragraph{Open-source MLLMs} While optimizing the sliding window parameters mitigates the expense of using commercial APIs, a persistent monetary cost remains for each evaluation. To address this fundamental challenge of cost and accessibility, we propose a lightweight, open-source alternative that eliminates API fees and allows for local deployment, thereby providing researchers with greater control and transparency. To this end, we introduce the A3RM (Android Agent Arena Reward Model), a specialized reward model fine-tuned from Qwen3-VL-8B \citep{bai2025qwen3vltechnicalreport}. We selected this base model due to its strong demonstrated capabilities in GUI related tasks, making it an ideal foundation for our evaluator. In summary, \textbf{A3RM} serves as an accurate, cost-effective, and transparent judge for essential-state verification.

\subsection{A3 Pipeline Suite}

To accompany our benchmark, we introduce a comprehensive, open-source software pipeline, as illustrated in Figure~\ref{fig:pipeline_suite}. The core of this suite is the AITK (Android Interaction Toolkit), a lightweight and customizable framework for managing agent-device interaction, data collection, and task execution. AITK is mainly designed for model-based agents and it features a plug-in architecture for integrating custom agents and tasks. Another component of the suite is evaluator, which implements our essential-state evaluation methods. This evaluator is designed to be agent-agnostic, allowing it to be applied to any standard trajectory data, independent of the agent and framework that produced it. The complete pipeline, including our human annotation tools, will be open-sourced to foster reproducibility and advance future research.

\section{Experiments}

\subsection{A3 Reward Model (A3RM)}

We developed the A3RM (Android Agent Arena Reward Model) as a deployable alternative reward model. We selected Qwen3-VL-8B \citep{bai2025qwen3vltechnicalreport} as the base model, given its strong performance on GUI related tasks. Our preliminary experiments revealed that a small window size of 2 yielded the best performance for the 8B model. This is likely due to its more limited context capacity. Therefore, we adopted a window size of 2 and interval size of 1 for all A3RM data collection and training.

\paragraph{Data Construction} We constructed a comprehensive training dataset by collecting an average of three distinct, successful human trajectories for each of the 100 benchmark tasks, ensuring the coverage of diverse valid trajectories. Human annotators manually identified the specific state transitions where essential states were achieved; these constitute our positive samples, while all other transitions within these expert trajectories serve as negative samples. To further improve the model's discriminative capability, we employed a negative sample mining strategy. Leveraging the high-recall, low-precision characteristic (Table~\ref{tab:a3rm}) of the Gemini-based evaluator, we processed agent-generated trajectories and labeled additional negative samples. The final dataset comprises 8241 step-wise samples derived from expert trajectories (981 positive and 7260 negative samples), augmented with 1083 negative samples from agent trajectories. 

\paragraph{Training} We finetuned Qwen3-VL-8B using the dynamic sampling policy optimization (DAPO) algorithm \citep{liu2025dapo} on our curated dataset. Given the natural class imbalance in our data (far more negative samples than positive ones), we oversampled the positive class by a factor of four in each training epoch to prevent the model from collapsing to a trivial solution of always predicting a negative outcome. 

\begin{table}[t]
    \centering
    \resizebox{0.99\linewidth}{!}{
        \begin{tabular}{l C{3cm} C{2.5cm}}
        \toprule
        \textbf{\quad Metric} & \textbf{Gemini-2.5-pro} & \textbf{A3RM} \\
        \midrule
        \multicolumn{3}{l}{\textit{Essential State Level}} \\
        \quad Precision  & 87.3 & 95.7 \\
        \quad Recall     & 96.3 & 94.8 \\
        \quad F1 Score   & 91.5 & 95.3 \\
        \quad Accuracy   & 89.5 & 96.6 \\
        \midrule
        \multicolumn{3}{l}{\textit{Task Level}} \\
        \quad Precision  & 85.7 & 96.0 \\
        \quad Recall     & 96.0 & 96.0 \\
        \quad F1 Score   & 90.6 & 96.0 \\
        \quad Accuracy   & 95.0 & 98.0 \\
        \bottomrule
        \end{tabular}
    }
    \caption{A3RM evaluation metric performance across different models. Metrics are reported for both Essential State (Ess. State) and Task levels.}
    \label{tab:a3rm}
\end{table}

\paragraph{Evaluator Performance} Table~\ref{tab:a3rm} presents the performance of our fine-tuned A3RM compared to the Gemini-2.5-pro baseline on a held-out A3 test set. Despite its significantly smaller size, A3RM surpasses the commercial model across key metrics. At the essential state level, it achieves 95.7\% Precision (+8.4 points), and at the task level, it dominates with 96.0\% Precision (+10.3 points) while maintaining parity in Recall. This performance validates our data construction strategy, which conditions the model to reject ambiguous states that zero-shot models frequently hallucinate as successes. However, we emphasize that A3RM is a specialized model optimized for the A3 task distribution; while this specialization yields superior in-domain performance compared to the general-purpose Gemini, it is designed specifically for this benchmark. Ultimately, this high precision is critical for reward modeling, as it minimizes the risk of inadvertently reinforcing agents for failed trajectories.

\begin{table*}[t]
    \centering
    \resizebox{0.98\linewidth}{!}{
    \begin{tabular}{l| C{1.7cm} | C{1.7cm} C{1.7cm} C{1.7cm} | C{2cm} C{2cm} | C{1.7cm}}
    \toprule
        \textbf{Agent} & \textbf{Metric} & \textbf{Easy} & \textbf{Medium} & \textbf{Hard} & \textbf{Operation} & \textbf{Inf. Query} & \textbf{Overall} \\
        \midrule \midrule
        \multirow{2}{*}{UI-TARS-1.5} & \cellcolor{gray!20}SR & \cellcolor{gray!20}20.0 & \cellcolor{gray!20}10.0 & \cellcolor{gray!20}4.0 & \cellcolor{gray!20}13.9 & \cellcolor{gray!20}7.1 & \cellcolor{gray!20}12 \\
         & ESAR & 35.3 & 21.8 & 23.9 & 31.2 & 20.9 & 28.2 \\
        \midrule

        \multirow{2}{*}{UI-Venus} & \cellcolor{gray!20}SR & \cellcolor{gray!20}28.5 & \cellcolor{gray!20}15.0 & \cellcolor{gray!20}16.0 & \cellcolor{gray!20}20.8 & \cellcolor{gray!20}17.8 & \cellcolor{gray!20}20 \\
         & ESAR & 41.2 & 29.7 & 27.1 & 33.9 & 27.5 & 32.0 \\
        \midrule

        \multirow{2}{*}{UI-Genie} & \cellcolor{gray!20}SR & \cellcolor{gray!20}25.8 & \cellcolor{gray!20}10.0 & \cellcolor{gray!20}0.0 & \cellcolor{gray!20}16.7 & \cellcolor{gray!20}3.6 & \cellcolor{gray!20}13 \\
         & ESAR & 41.2 & 25.8 & 32.3 & 34.9 & 25.3 & 32.1 \\
        \midrule

        \multirow{2}{*}{InfiGUI-R1} & \cellcolor{gray!20}SR & \cellcolor{gray!20}34.3 & \cellcolor{gray!20}30.0 & \cellcolor{gray!20}12.0 & \cellcolor{gray!20}34.7 & \cellcolor{gray!20}7.1 & \cellcolor{gray!20}27 \\
         & ESAR & 55.3 & 50.8 & 51.0 & 54.6 & 46.2 & 52.1 \\
        \midrule

        \multirow{2}{*}{GUI-OWL} & \cellcolor{gray!20}SR & \cellcolor{gray!20}31.4 & \cellcolor{gray!20}7.5 & \cellcolor{gray!20}0.0 & \cellcolor{gray!20}18.1 & \cellcolor{gray!20}3.6 & \cellcolor{gray!20}14 \\
         & ESAR & 49.4 & 26.6 & 23.9 & 35.8 & 23.1 & 32.0 \\
        \midrule

        \multirow{2}{*}{Qwen2.5-VL} & \cellcolor{gray!20}SR & \cellcolor{gray!20}5.7 & \cellcolor{gray!20}0.0 & \cellcolor{gray!20}4.0 & \cellcolor{gray!20}4.2 & \cellcolor{gray!20}0.0 & \cellcolor{gray!20}3 \\
         & ESAR & 10.5 & 10.9 & 21.8 & 15.6 & 11.0 & 14.2 \\
        \midrule

        \multirow{2}{*}{Qwen3-VL} & \cellcolor{gray!20}SR & \cellcolor{gray!20}34.3 & \cellcolor{gray!20}12.5 & \cellcolor{gray!20}0.0 & \cellcolor{gray!20}20.8 & \cellcolor{gray!20}7.1 & \cellcolor{gray!20}17 \\
         & ESAR & 50.6 & 35.2 & 31.3 & 44.0 & 24.2 & 38.2 \\
        \midrule \midrule

        \multirow{2}{*}{\makecell[l]{Mobile-Use + \\ Qwen2.5-VL}} & \cellcolor{gray!20}SR & \cellcolor{gray!20}34.3 & \cellcolor{gray!20}10.0 & \cellcolor{gray!20}0.0 & \cellcolor{gray!20}22.2 & \cellcolor{gray!20}0.0 & \cellcolor{gray!20}16 \\
         & ESAR & 48.2 & 39.9 & 31.3 & 43.1 & 30.8 & 39.5 \\
        \midrule

        \multirow{2}{*}{\makecell[l]{T3A + \\ Qwen2.5-VL}} & \cellcolor{gray!20}SR & \cellcolor{gray!20}31.4 & \cellcolor{gray!20}10.0 & \cellcolor{gray!20}4.0 & \cellcolor{gray!20}19.4 & \cellcolor{gray!20}3.8 & \cellcolor{gray!20}15 \\
         & ESAR & 37.6 & 28.1 & 28.1 & 29.4 & 34.1 & 30.7 \\
        \midrule

        \multirow{2}{*}{\makecell[l]{T3A + \\ Gemini-2.5-pro}} & \cellcolor{gray!20}SR & \cellcolor{gray!20}57.1 & \cellcolor{gray!20}55.0 & \cellcolor{gray!20}44.0 & \cellcolor{gray!20}55.6 & \cellcolor{gray!20}46.6 & \cellcolor{gray!20}53 \\
         & ESAR & 68.2 & 67.9 & 62.5 & 72.9 & 50.5 & 66.4 \\

    \bottomrule
    \end{tabular}
    }
    \caption{Benchmark results evaluated by our A3RM. We report Task Success Rate (SR) and Essential State Achieved Rate (ESAR) across task categories and difficulties.}
    \label{tab:a3_a3rm}
\end{table*}

\subsection{A3 Benchmark}

\paragraph{Experimental Setup} To establish a comprehensive baseline, we evaluated a diverse suite of mobile GUI agents. We selected seven representative single-model agents: Qwen2.5-VL-7B \citep{bai2025qwen25vltechnicalreport}, Qwen3-VL-8B~\citep{bai2025qwen3vltechnicalreport}, UI-TARS-1.5-7B \citep{qin2025ui-tars}, UI-Genie-7B \citep{xiao2025uigenie}, UI-Venus-7B \citep{gu2025uivenustechnicalreportbuilding}, InfiGUI-R1-3B \citep{liu2025infiguir1advancingmultimodalgui}, and GUI-OWL-7B \citep{ye2025mobileagentv3fundamentalagentsgui}. Additionally, we evaluated two agent frameworks: Mobile-Use~\citep{li2025mobileuse} (powered by Qwen2.5-VL-7B) and T3A~\citep{rawlesandroidworld} (evaluated with both Gemini-2.5-pro and Qwen2.5-VL-7B). All agents were evaluated on the A3 benchmark using their official, publicly available prompts and inference settings to ensure reproducibility. Detailed results are presented in Table~\ref{tab:a3_a3rm} (A3RM evaluation) and Table~\ref{tab:a3_gemini} (Gemini evaluation in Appendix~\ref{app:gemini}).

\paragraph{Performance Analysis} The results indicate that A3 poses a rigorous challenge for current state-of-the-art agents. Among open-source single-model agents, InfiGUI-R1 distinguishes itself as the leader with a Success Rate (SR) of 27.0\%. However, the broader landscape reveals significant fragility, where most models exhibit precipitous performance declines on "Hard" tasks, where success rates frequently approach zero. In sharp contrast, the proprietary T3A + Gemini-2.5-pro agent establishes a strong upper bound with a 53.0\% SR. This nearly two-fold performance gap underscores a critical reality: while specialized open-source agents are evolving, they still lag significantly behind large-scale commercial MLLMs in the complex, multi-step reasoning required for dynamic GUIs. Ultimately, with even the top-performing system failing nearly half the tasks, A3 reveals substantial room for improvement across the board, positioning the commercial model's performance as a current "skyline" for the open-source community to pursue.

\paragraph{Frameworks vs. Foundation Models} Our ablation of frameworks versus base models reveals critical insights. First, structured frameworks significantly enhance weak base models: Qwen2.5-VL fails almost completely as a standalone agent (3.0\% SR), but when integrated into the Mobile-Use or T3A frameworks, its performance jumps to 16.0\% and 15.0\% respectively. This suggests that well designed frameworks can compensate for limited reasoning capabilities. However, the backbone model remains the upper bound of performance. When the exact same T3A framework is powered by Gemini-2.5-pro, performance skyrockets to 53.0\% (compared to 15.0\% with Qwen2.5). This demonstrates that while frameworks provide necessary structure, they cannot fully mitigate the reasoning deficits of the underlying vision-language model. We also observe rapid evolution in base models; Qwen3-VL (17.0\% SR) naturally outperforms its predecessor Qwen2.5-VL (3.0\% SR) by a wide margin without any framework assistance.

\paragraph{Essential State Evaluation} The Essential State Achieved Rate (ESAR) provides a more granular view of agent capabilities than the binary SR metric. A consistent trend across all agents is that ESAR is substantially higher than SR. This discrepancy highlights the primary failure mode of current agents: they possess sufficient semantic understanding to initiate tasks and navigate early stages (high ESAR) but lack the long-horizon robustness required to reach the final state without encountering a terminal error (low SR). This phenomenon also corresponds to our case studies in Appendix~\ref{app:case}, where current agents mostly fail in progress awareness.

We also provide more A3RM and essential state evaluation generalization in Appendix~\ref{app:generalization}, sliding window analysis in Appendix~\ref{app:sliding_window} and Gemini evaluation experiments and results in Appendix~\ref{app:gemini}.

\section{Conclusion}

We introduced the Android Agent Arena (A3), a benchmark that bridges the gap between static evaluations and real-world utility by incorporating dynamic tasks from popular online applications. To overcome the opacity of closed-source apps, we proposed the essential-state evaluation method and the fine-tuned A3RM, offering a scalable, visual-based metric that captures granular agent progress beyond binary success. Our experiments reveal that while current agents struggle with long-horizon robustness, our open-sourced pipeline provides the necessary infrastructure to accelerate the development of truly autonomous mobile assistants.

\section*{Limitations}

First, our benchmark is confined to the Android ecosystem. Expanding to iOS remains effectively unsolvable due to restrictive system policies that prevent the virtualization and programmatic control necessary for reproducible testing. Second, despite our rigorous curation, we exclude specific high-frequency categories like instant messaging, where strict authentication protocols and privacy concerns make automated account resets impractical. Furthermore, while our A3RM evaluator demonstrates high precision, it is inherently a probabilistic model and still has hallucinations. Consequently, while A3 significantly advances ecological validity, these constraints highlight the trade-offs required to balance reproducibility with the closed, secure nature of modern mobile platforms.

\bibliography{custom}

@article{rawles2023androidinthewild,
  title={Androidinthewild: A large-scale dataset for android device control},
  author={Rawles, Christopher and Li, Alice and Rodriguez, Daniel and Riva, Oriana and Lillicrap, Timothy},
  journal={Advances in Neural Information Processing Systems},
  volume={36},
  pages={59708--59728},
  year={2023}
}

@inproceedings{chai-etal-2025-amex,
    title = "{AMEX}: Android Multi-annotation Expo Dataset for Mobile {GUI} Agents",
    author = "Chai, Yuxiang  and
      Huang, Siyuan  and
      Niu, Yazhe  and
      Xiao, Han  and
      Liu, Liang  and
      Wang, Guozhi  and
      Zhang, Dingyu  and
      Ren, Shuai  and
      Li, Hongsheng",
    editor = "Che, Wanxiang  and
      Nabende, Joyce  and
      Shutova, Ekaterina  and
      Pilehvar, Mohammad Taher",
    booktitle = "Findings of the Association for Computational Linguistics: ACL 2025",
    month = jul,
    year = "2025",
    address = "Vienna, Austria",
    publisher = "Association for Computational Linguistics",
    url = "https://aclanthology.org/2025.findings-acl.110/",
    doi = "10.18653/v1/2025.findings-acl.110",
    pages = "2138--2156",
    ISBN = "979-8-89176-256-5",
}

@article{li2024effects,
  title={On the effects of data scale on ui control agents},
  author={Li, Wei and Bishop, William E and Li, Alice and Rawles, Christopher and Campbell-Ajala, Folawiyo and Tyamagundlu, Divya and Riva, Oriana},
  journal={Advances in Neural Information Processing Systems},
  volume={37},
  pages={92130--92154},
  year={2024}
}

@inproceedings{rawlesandroidworld,
  title={AndroidWorld: A Dynamic Benchmarking Environment for Autonomous Agents},
  author={Rawles, Christopher and Clinckemaillie, Sarah and Chang, Yifan and Waltz, Jonathan and Lau, Gabrielle and Fair, Marybeth and Li, Alice and Bishop, William E and Li, Wei and Campbell-Ajala, Folawiyo and others},
  year={2025},
  booktitle={The Thirteenth International Conference on Learning Representations}
}

@inproceedings{chen2024spa,
  title={Spa-bench: A comprehensive benchmark for smartphone agent evaluation},
  author={Chen, Jingxuan and Yuen, Derek and Xie, Bin and Yang, Yuhao and Chen, Gongwei and Wu, Zhihao and Yixing, Li and Zhou, Xurui and Liu, Weiwen and Wang, Shuai and others},
  booktitle={NeurIPS 2024 Workshop on Open-World Agents},
  year={2024}
}

@inproceedings{xing2024androidarena,
  title={Understanding the weakness of large language model agents within a complex android environment},
  author={Xing, Mingzhe and Zhang, Rongkai and Xue, Hui and Chen, Qi and Yang, Fan and Xiao, Zhen},
  booktitle={Proceedings of the 30th ACM SIGKDD Conference on Knowledge Discovery and Data Mining},
  pages={6061--6072},
  year={2024}
}

@inproceedings{xu-etal-2025-androidlab,
    title = "{A}ndroid{L}ab: Training and Systematic Benchmarking of Android Autonomous Agents",
    author = "Xu, Yifan  and
      Liu, Xiao  and
      Sun, Xueqiao  and
      Cheng, Siyi  and
      Yu, Hao  and
      Lai, Hanyu  and
      Zhang, Shudan  and
      Zhang, Dan  and
      Tang, Jie  and
      Dong, Yuxiao",
    editor = "Che, Wanxiang  and
      Nabende, Joyce  and
      Shutova, Ekaterina  and
      Pilehvar, Mohammad Taher",
    booktitle = "Proceedings of the 63rd Annual Meeting of the Association for Computational Linguistics (Volume 1: Long Papers)",
    month = jul,
    year = "2025",
    address = "Vienna, Austria",
    publisher = "Association for Computational Linguistics",
    url = "https://aclanthology.org/2025.acl-long.107/",
    doi = "10.18653/v1/2025.acl-long.107",
    pages = "2144--2166",
    ISBN = "979-8-89176-251-0"
}

@article{zhang2023mobileenv,
  title={Mobile-env: an evaluation platform and benchmark for LLM-GUI interaction},
  author={Zhang, Danyang and Xu, Hongshen and Zhao, Zihan and Chen, Lu and Cao, Ruisheng and Yu, Kai},
  journal={arXiv preprint arXiv:2305.08144},
  year={2023}
}

@article{qin2025ui-tars,
  title={UI-TARS: Pioneering Automated GUI Interaction with Native Agents},
  author={Qin, Yujia and Ye, Yining and Fang, Junjie and Wang, Haoming and Liang, Shihao and Tian, Shizuo and Zhang, Junda and Li, Jiahao and Li, Yunxin and Huang, Shijue and others},
  journal={arXiv preprint arXiv:2501.12326},
  year={2025}
}

@article{
liu2025llmpowered,
title={{LLM}-Powered {GUI} Agents in Phone Automation: Surveying Progress and Prospects},
author={Guangyi Liu and Pengxiang Zhao and Yaozhen Liang and Liang Liu and Yaxuan Guo and Han Xiao and Weifeng Lin and Yuxiang Chai and Yue Han and Shuai Ren and Hao Wang and Xiaoyu Liang and WenHao Wang and Tianze Wu and Zhengxi Lu and Siheng Chen and LiLinghao and Hao Wang and Guanjing Xiong and Yong Liu and Hongsheng Li},
journal={Transactions on Machine Learning Research},
issn={2835-8856},
year={2025},
url={https://openreview.net/forum?id=yWQqoi1G1K},
note={}
}

@misc{chen2024mllmasajudgeassessingmultimodalllmasajudge,
      title={MLLM-as-a-Judge: Assessing Multimodal LLM-as-a-Judge with Vision-Language Benchmark}, 
      author={Dongping Chen and Ruoxi Chen and Shilin Zhang and Yinuo Liu and Yaochen Wang and Huichi Zhou and Qihui Zhang and Yao Wan and Pan Zhou and Lichao Sun},
      year={2024},
      eprint={2402.04788},
      archivePrefix={arXiv},
      primaryClass={cs.CL},
      url={https://arxiv.org/abs/2402.04788}, 
}

@misc{bai2025qwen3vltechnicalreport,
      title={Qwen3-VL Technical Report}, 
      author={Shuai Bai and Yuxuan Cai and Ruizhe Chen and Keqin Chen and Xionghui Chen and Zesen Cheng and Lianghao Deng and Wei Ding and Chang Gao and Chunjiang Ge and Wenbin Ge and Zhifang Guo and Qidong Huang and Jie Huang and Fei Huang and Binyuan Hui and Shutong Jiang and Zhaohai Li and Mingsheng Li and Mei Li and Kaixin Li and Zicheng Lin and Junyang Lin and Xuejing Liu and Jiawei Liu and Chenglong Liu and Yang Liu and Dayiheng Liu and Shixuan Liu and Dunjie Lu and Ruilin Luo and Chenxu Lv and Rui Men and Lingchen Meng and Xuancheng Ren and Xingzhang Ren and Sibo Song and Yuchong Sun and Jun Tang and Jianhong Tu and Jianqiang Wan and Peng Wang and Pengfei Wang and Qiuyue Wang and Yuxuan Wang and Tianbao Xie and Yiheng Xu and Haiyang Xu and Jin Xu and Zhibo Yang and Mingkun Yang and Jianxin Yang and An Yang and Bowen Yu and Fei Zhang and Hang Zhang and Xi Zhang and Bo Zheng and Humen Zhong and Jingren Zhou and Fan Zhou and Jing Zhou and Yuanzhi Zhu and Ke Zhu},
      year={2025},
      eprint={2511.21631},
      archivePrefix={arXiv},
      primaryClass={cs.CV},
      url={https://arxiv.org/abs/2511.21631}, 
}

@misc{wang2025guiagentsfoundationmodels,
      title={GUI Agents with Foundation Models: A Comprehensive Survey}, 
      author={Shuai Wang and Weiwen Liu and Jingxuan Chen and Yuqi Zhou and Weinan Gan and Xingshan Zeng and Yuhan Che and Shuai Yu and Xinlong Hao and Kun Shao and Bin Wang and Chuhan Wu and Yasheng Wang and Ruiming Tang and Jianye Hao},
      year={2025},
      eprint={2411.04890},
      archivePrefix={arXiv},
      primaryClass={cs.AI},
      url={https://arxiv.org/abs/2411.04890}, 
}

@misc{ye2025mobileagentv3fundamentalagentsgui,
      title={Mobile-Agent-v3: Fundamental Agents for GUI Automation}, 
      author={Jiabo Ye and Xi Zhang and Haiyang Xu and Haowei Liu and Junyang Wang and Zhaoqing Zhu and Ziwei Zheng and Feiyu Gao and Junjie Cao and Zhengxi Lu and Jitong Liao and Qi Zheng and Fei Huang and Jingren Zhou and Ming Yan},
      year={2025},
      eprint={2508.15144},
      archivePrefix={arXiv},
      primaryClass={cs.AI},
      url={https://arxiv.org/abs/2508.15144}, 
}

@inproceedings{hu-etal-2025-os,
    title = "{OS} Agents: A Survey on {MLLM}-based Agents for Computer, Phone and Browser Use",
    author = "Hu, Xueyu  and
      Xiong, Tao  and
      Yi, Biao  and
      Wei, Zishu  and
      Xiao, Ruixuan  and
      Chen, Yurun  and
      Ye, Jiasheng  and
      Tao, Meiling  and
      Zhou, Xiangxin  and
      Zhao, Ziyu  and
      Li, Yuhuai  and
      Xu, Shengze  and
      Wang, Shenzhi  and
      Xu, Xinchen  and
      Qiao, Shuofei  and
      Wang, Zhaokai  and
      Kuang, Kun  and
      Zeng, Tieyong  and
      Wang, Liang  and
      Li, Jiwei  and
      Jiang, Yuchen Eleanor  and
      Zhou, Wangchunshu  and
      Wang, Guoyin  and
      Yin, Keting  and
      Zhao, Zhou  and
      Yang, Hongxia  and
      Wu, Fan  and
      Zhang, Shengyu  and
      Wu, Fei",
    editor = "Che, Wanxiang  and
      Nabende, Joyce  and
      Shutova, Ekaterina  and
      Pilehvar, Mohammad Taher",
    booktitle = "Proceedings of the 63rd Annual Meeting of the Association for Computational Linguistics (Volume 1: Long Papers)",
    month = jul,
    year = "2025",
    address = "Vienna, Austria",
    publisher = "Association for Computational Linguistics",
    url = "https://aclanthology.org/2025.acl-long.369/",
    doi = "10.18653/v1/2025.acl-long.369",
    pages = "7436--7465",
    ISBN = "979-8-89176-251-0",
}

@misc{bai2025qwen25vltechnicalreport,
      title={Qwen2.5-VL Technical Report}, 
      author={Shuai Bai and Keqin Chen and Xuejing Liu and Jialin Wang and Wenbin Ge and Sibo Song and Kai Dang and Peng Wang and Shijie Wang and Jun Tang and Humen Zhong and Yuanzhi Zhu and Mingkun Yang and Zhaohai Li and Jianqiang Wan and Pengfei Wang and Wei Ding and Zheren Fu and Yiheng Xu and Jiabo Ye and Xi Zhang and Tianbao Xie and Zesen Cheng and Hang Zhang and Zhibo Yang and Haiyang Xu and Junyang Lin},
      year={2025},
      eprint={2502.13923},
      archivePrefix={arXiv},
      primaryClass={cs.CV},
      url={https://arxiv.org/abs/2502.13923}, 
}

@inproceedings{
liu2025dapo,
title={{DAPO} : Improving Multi-Step Reasoning Abilities of Large Language Models with Direct Advantage-Based Policy Optimization},
author={Jiacai Liu and Chaojie Wang and Chris Yuhao Liu and Liang Zeng and Rui Yan and Yiwen Sun and Yang Liu},
booktitle={The Thirty-ninth Annual Conference on Neural Information Processing Systems},
year={2025},
url={https://openreview.net/forum?id=77eEDRhPkQ}
}

@inproceedings{
xiao2025uigenie,
title={{UI}-Genie: A Self-Improving Approach for Iteratively Boosting {MLLM}-based Mobile {GUI} Agents},
author={Han Xiao and Guozhi Wang and Yuxiang Chai and Zimu Lu and Weifeng Lin and Hao He and Lue Fan and Liuyang Bian and Rui Hu and Liang Liu and Shuai Ren and Yafei Wen and Xiaoxin Chen and Aojun Zhou and Hongsheng Li},
booktitle={The Thirty-ninth Annual Conference on Neural Information Processing Systems},
year={2025},
url={https://openreview.net/forum?id=3uUmJzSSOW}
}

@misc{gu2025uivenustechnicalreportbuilding,
      title={UI-Venus Technical Report: Building High-performance UI Agents with RFT}, 
      author={Zhangxuan Gu and Zhengwen Zeng and Zhenyu Xu and Xingran Zhou and Shuheng Shen and Yunfei Liu and Beitong Zhou and Changhua Meng and Tianyu Xia and Weizhi Chen and Yue Wen and Jingya Dou and Fei Tang and Jinzhen Lin and Yulin Liu and Zhenlin Guo and Yichen Gong and Heng Jia and Changlong Gao and Yuan Guo and Yong Deng and Zhenyu Guo and Liang Chen and Weiqiang Wang},
      year={2025},
      eprint={2508.10833},
      archivePrefix={arXiv},
      primaryClass={cs.CV},
      url={https://arxiv.org/abs/2508.10833}, 
}

@misc{liu2025infiguir1advancingmultimodalgui,
      title={InfiGUI-R1: Advancing Multimodal GUI Agents from Reactive Actors to Deliberative Reasoners}, 
      author={Yuhang Liu and Pengxiang Li and Congkai Xie and Xavier Hu and Xiaotian Han and Shengyu Zhang and Hongxia Yang and Fei Wu},
      year={2025},
      eprint={2504.14239},
      archivePrefix={arXiv},
      primaryClass={cs.AI},
      url={https://arxiv.org/abs/2504.14239}, 
}

@inproceedings{
li2025mobileuse,
title={MobileUse: A Hierarchical Reflection-Driven {GUI} Agent for Autonomous Mobile Operation},
author={Ning Li and Xiangmou Qu and Jiamu Zhou and Jun Wang and Muning Wen and Kounianhua Du and Xingyu Lou and Qiuying Peng and Jun Wang and Weinan Zhang},
booktitle={The Thirty-ninth Annual Conference on Neural Information Processing Systems},
year={2025},
url={https://openreview.net/forum?id=KR6tnkb6h4}
}

@misc{lee2025benchmarkingmobiledevicecontrol,
      title={Benchmarking Mobile Device Control Agents across Diverse Configurations}, 
      author={Juyong Lee and Taywon Min and Minyong An and Dongyoon Hahm and Haeone Lee and Changyeon Kim and Kimin Lee},
      year={2025},
      eprint={2404.16660},
      archivePrefix={arXiv},
      primaryClass={cs.HC},
      url={https://arxiv.org/abs/2404.16660}, 
}

@misc{zheng2025easyr1,
  title        = {EasyR1: An Efficient, Scalable, Multi-Modality RL Training Framework},
  author       = {Yaowei Zheng and Junting Lu and Shenzhi Wang and Zhangchi Feng and Dongdong Kuang and Yuwen Xiong},
  howpublished = {\url{https://github.com/hiyouga/EasyR1}},
  year         = {2025}
}

\clearpage

\appendix

\section{Appendix}

\subsection{App List}
\label{app:app_list}

We select 20 apps in 20 categories from Google Play Store top charts as listed in Table~\ref{tab:app_list}.

\begin{table}[h]
    \centering
    \resizebox{0.98\linewidth}{!}{
    \begin{tabular}{C{3.5cm} C{3.5cm}}
    \toprule
        \textbf{App} & \textbf{Category} \\
        \midrule
        CNN & News \\ \midrule TripAdvisor & Events \\  \midrule  Amazon & Shopping \\ \midrule 
        Omio & Travel \\ \midrule  Bluecoins & Finance \\ \midrule  Tasks & Productivity \\ \midrule 
        N Calendar & Calendar \\ \midrule  File & Tools \\ \midrule  eboox & Read \\ \midrule 
        YouTube & Video \\ \midrule  Home Workout & Fitness \\ \midrule  Pinterest & Lifestyle \\ \midrule 
        Chrome & Browser \\ \midrule  CityMapper & Navigation \\ \midrule  Joytify & Music \\ \midrule 
        Joplin & Notes \\ \midrule  Wikipedia & Entertainment \\ \midrule  Weather Forecast & Weather \\ \midrule 
        Supercook & Food \\ \midrule  Gmail & Business \\
    \bottomrule
    \end{tabular}
    }
    \caption{List of apps and corresponding categories.}
    \label{tab:app_list}
\end{table}

\subsection{Tasks \& Essential States}
\label{app:tasks}

For clearer demonstration of our tasks and essential states, we list 10 examples in Table~\ref{tab:task_examples}.

\subsection{Sliding Window and Interval Study}
\label{app:sliding_window}

We conduct experiments on the sliding window size and interval size (number of steps) for the commercial MLLMs (Gemini) evaluation to identify the optimal parameters to minimize the number of API calls without compromising evaluation fidelity. Table~\ref{tab:window} summarizes the evaluation results across different sliding window sizes and interval settings. Specifically, the sliding window size ranges from 2 to 6, with interval sizes set to half of the window size. From the results, we observe that sliding window sizes of 3 and 4 yield the highest evaluation accuracies for both essential states and overall task performance. Among these, the configuration with a sliding window size of 4 and an interval size of 2 achieves the lowest average computational cost, making it the preferred setting for the evaluation method. 

Further analysis of our results reveals a distinct trade-off in the selection of the sliding window size, with the MLLM evaluator's performance degrading at both small and large extremes. When a small window is combined with a non-overlapping stride (e.g., size of 2, stride of 2), evaluation accuracy suffers. This occurs because a critical state transition can be split across the boundary of two consecutive windows. For example, an agent's action and its immediate on-screen result might fall into separate windows, depriving the MLLM of the necessary context to make a correct judgment. Conversely, when the window size becomes too large (e.g., greater than 4), performance degrades due to loss of visual fidelity. To be processed, frames in the window are concatenated into a single composite image. With larger window sizes, each frame is significantly downscaled, causing a severe loss of resolution that can render small text and UI elements illegible. This image compression impairs the MLLM's ability to accurately identify essential states, leading to misjudgments.

\begin{table*}[t]
    \centering
    \resizebox{0.98\linewidth}{!}{
    \begin{tabular}{C{2.5cm} C{2.5cm} C{3.2cm} C{2.5cm} C{3cm}}
        \toprule
        \textbf{Window Size} & \textbf{Interval Size} & \textbf{Essential-State Acc} & \textbf{Task Acc} & \textbf{Average Cost} (\$)\\
        \midrule
        \multirow{2}{*}{2} & 1 & 0.94 & \textbf{0.91} & 0.196 \\
         & 2 & 0.88 & 0.86 & 0.099 \\
        \midrule
        \multirow{2}{*}{3} & 1 & \textbf{0.95} & \textbf{0.91} & 0.195 \\
         & 2 & 0.94 & 0.90 & 0.098 \\
        \midrule
        \multirow{2}{*}{4} & 2 & \textbf{0.95} & \textbf{0.91} & 0.098 \\
         & 3 & \textbf{0.95} & 0.90 & 0.071 \\
        \midrule
        \multirow{2}{*}{5} & 2 & 0.93 & 0.89 & 0.097 \\
         & 3 & 0.92 & 0.89 & 0.069 \\
        \midrule
        \multirow{2}{*}{6} & 3 & 0.91 & 0.88 & 0.070 \\
         & 4 & 0.89 & 0.87 & 0.050 \\
        \bottomrule
    \end{tabular}
    }
    \caption{Sliding window and interval size study for essential-state evaluation. The highest accuracy is in bold. The average cost is computed by the total API cost of three MLLMs over 30 tasks.}
    \label{tab:window}
\end{table*}

\subsection{Gemini Evaluation}
\label{app:gemini}

\paragraph{Gemini-Based Evaluation Analysis} The results from the Gemini-based evaluation (Table~\ref{tab:a3_gemini}) highlight the critical importance of foundational model capability and agentic scaffolding. The proprietary T3A + Gemini-2.5-pro agent establishes a dominant upper bound with an Overall Success Rate (SR) of 58.0\%, significantly outperforming all open-source alternatives. Among single-model agents, InfiGUI-R1 leads with a 29.0\% SR, demonstrating that specialized fine-tuning can yield competitive results. We also observe a substantial generational leap in the Qwen family: Qwen3-VL achieves a 23.0\% SR, a nearly five-fold improvement over the baseline Qwen2.5-VL (5.0\%). Furthermore, the data underscores the efficacy of agent frameworks; wrapping the weak Qwen2.5-VL base model in the T3A framework boosts its performance from 5.0\% to 24.0\%, effectively bridging the gap to the stronger Qwen3-VL base model. However, despite these gains, the sharp decline in performance on "Hard" tasks—where most open-source agents drop to single-digit success rates—indicates that complex, multi-step reasoning remains a significant bottleneck.

\paragraph{Gemini VS A3RM} Comparing these results with the A3RM evaluation reveals a systematic "optimism bias" in the commercial Gemini evaluator. Across virtually all agents, Gemini assigns higher success rates than our fine-tuned A3RM. For instance, the leading T3A + Gemini agent is scored at 58.0\% by the Gemini evaluator but drops to 53.0\% under A3RM. Similarly, InfiGUI-R1 decreases from 29.0\% to 27.0\%, and Mobile-Use drops from 19.0\% to 16.0\%. We attribute this discrepancy to the design of A3RM: explicitly trained with human labels and hard negative mining, A3RM is more aligned to human judgment and Gemini suffers from the low precision situation. This suggests that while commercial MLLMs offer high recall, they lack the domain-specific precision required for rigorous GUI evaluation. Crucially, however, the relative rankings of the agents remain largely consistent across both evaluators (e.g., T3A > InfiGUI > Qwen2.5).

\begin{table*}[t]
    \centering
    \resizebox{0.98\linewidth}{!}{
    \begin{tabular}{l| c | C{1.7cm} C{1.7cm} C{1.7cm} | C{2cm} C{2cm} | C{1.7cm}}
    \toprule
        \textbf{Agent} & \textbf{Metric} & \textbf{Easy} & \textbf{Medium} & \textbf{Hard} & \textbf{Operation} & \textbf{Inf. Query} & \textbf{Overall} \\
        \midrule \midrule
        \multirow{2}{*}{UI-TARS-1.5} & \cellcolor{gray!20}SR & \cellcolor{gray!20}14.3 & \cellcolor{gray!20}5.0 & \cellcolor{gray!20}4.0 & \cellcolor{gray!20}9.7 & \cellcolor{gray!20}3.5 & \cellcolor{gray!20}8 \\
         & ESAR & 29.4 & 17.2 & 9.3 & 20.2 & 13.2 & 18.1  \\
        \midrule
        
        \multirow{2}{*}{UI-Venus} & \cellcolor{gray!20}SR & \cellcolor{gray!20}31.4 & \cellcolor{gray!20}17.5 & \cellcolor{gray!20}12.0 & \cellcolor{gray!20}22.2 & \cellcolor{gray!20}17.8 & \cellcolor{gray!20}21 \\
         & ESAR & 47.1 & 27.3 & 31.2 & 36.2 & 28.6 & 34.0 \\
        \midrule
        
        \multirow{2}{*}{UI-Genie} & \cellcolor{gray!20}SR & \cellcolor{gray!20}28.6 & \cellcolor{gray!20}15.0 & \cellcolor{gray!20}0.0 & \cellcolor{gray!20}20.8 & \cellcolor{gray!20}3.6 & \cellcolor{gray!20}16 \\
         & ESAR & 47.1 & 28.1 & 31.3 & 37.2 & 27.5 & 34.3 \\
        \midrule
        
        \multirow{2}{*}{InfiGUI-R1} & \cellcolor{gray!20}SR & \cellcolor{gray!20}40.0 & \cellcolor{gray!20}32.5 & \cellcolor{gray!20}8.0 & \cellcolor{gray!20}36.1 & \cellcolor{gray!20}10.7 & \cellcolor{gray!20}29\\
         & ESAR & 58.8 & 53.1 & 52.1 & 56.4 & 49.5 & 54.4 \\
        \midrule
        
        \multirow{2}{*}{GUI-OWL} & \cellcolor{gray!20}SR & \cellcolor{gray!20}25.7 & \cellcolor{gray!20}5.0 & \cellcolor{gray!20}4.0 & \cellcolor{gray!20}15.3 & \cellcolor{gray!20}3.6 & \cellcolor{gray!20}12 \\
         & ESAR & 51.7 & 27.3 & 21.8 & 36.2 & 23.1 & 32.4 \\
         \midrule
        
        \multirow{2}{*}{Qwen2.5-VL} & \cellcolor{gray!20}SR & \cellcolor{gray!20}8.6 & \cellcolor{gray!20}2.5 & \cellcolor{gray!20}4.0 & \cellcolor{gray!20}6.9 & \cellcolor{gray!20}0.0 & \cellcolor{gray!20}5   \\
         & ESAR & 25.9 & 14.8 & 26.0 & 22.0 & 19.8 & 21.4\\
        \midrule
        
        \multirow{2}{*}{Qwen3-VL} & \cellcolor{gray!20}SR & \cellcolor{gray!20}40.0 & \cellcolor{gray!20}17.5 & \cellcolor{gray!20}8.0 & \cellcolor{gray!20}25.0 & \cellcolor{gray!20}17.9 & \cellcolor{gray!20}23   \\
         & ESAR & 55.3 & 40.6 & 36.5 & 50.5 & 37.4 & 46.7 \\
        
        \midrule \midrule
        
        \multirow{2}{*}{\makecell[l]{Mobile-Use + \\ Qwen2.5-VL}} & \cellcolor{gray!20}SR & \cellcolor{gray!20}34.3 & \cellcolor{gray!20}15.0 & \cellcolor{gray!20}4.0 & \cellcolor{gray!20}25.0 & \cellcolor{gray!20}3.6 & \cellcolor{gray!20}19 \\
         & ESAR & 47.1 & 43.0 & 30.2 & 43.1 & 33.0 & 40.1 \\
        \midrule

        \multirow{2}{*}{\makecell[l]{T3A + \\ Qwen2.5-VL}} & \cellcolor{gray!20}SR & \cellcolor{gray!20}40.0 & \cellcolor{gray!20}17.5 & \cellcolor{gray!20}12.0 & \cellcolor{gray!20}26.4 & \cellcolor{gray!20}17.9 & \cellcolor{gray!20}24 \\
         & ESAR & 44.7 & 32.8 & 31.3 & 32.1 & 43.9 & 35.6 \\
        \midrule
        
        \multirow{2}{*}{\makecell[l]{T3A + \\ Gemini-2.5-pro}} & \cellcolor{gray!20}SR & \cellcolor{gray!20}62.8 & \cellcolor{gray!20}57.5 & \cellcolor{gray!20}52.0 & \cellcolor{gray!20}62.5 & \cellcolor{gray!20}46.4 & \cellcolor{gray!20}58 \\
         & ESAR & 71.76 & 70.3 & 65.6 & 76.6 & 51.6 & 69.3 \\

    \bottomrule
    \end{tabular}
    }
    \caption{Benchmark results evaluated by the commercial Gemini-2.5-pro model. We report Task Success Rate (SR) and Essential State Achieved Rate (ESAR) across task categories and difficulties.}
    \label{tab:a3_gemini}
\end{table*}

\subsection{Case Study}
\label{app:case}

Through observation and analysis of the ten agents performance on A3 benchmark, we noticed some representative and remarkable cases where researchers would like to solve in the future study.

\subsubsection{General Agent Issues}

\begin{figure*}
    \centering
    \includegraphics[width=0.9\linewidth]{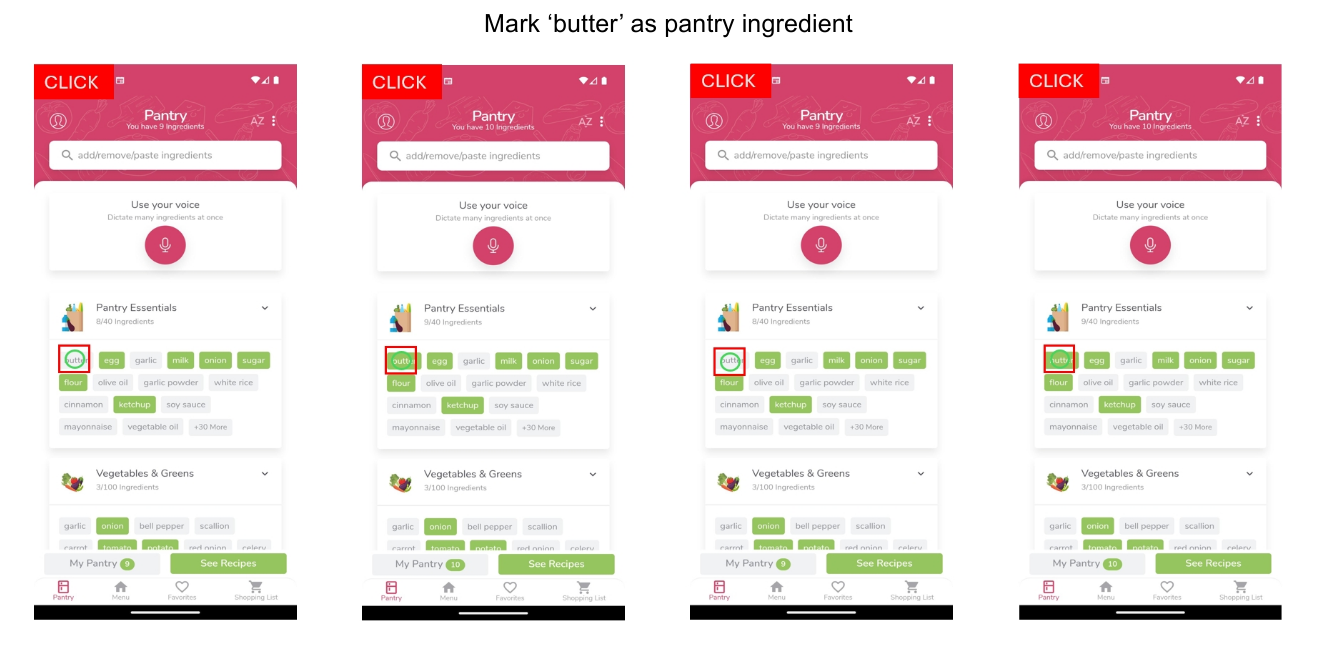}
    \caption{Example for redundant actions in an essential state.}
    \label{fig:bad_case_1}
\end{figure*}

\paragraph{Progress Unawareness} A critical failure mode observed across most model-based agents is Progress Unawareness, a fundamental inability to track their position within a task sequence. This limitation typically manifests in two ways:
\begin{itemize}[leftmargin=5mm]
    \item \textbf{Redundant Actions}: Agents often fail to recognize that a required step has already been completed. We observed numerous instances where an agent would get stuck in a loop, repeatedly attempting to achieve an essential state that was already satisfied, even when its action history was provided in the prompt (Figure~\ref{fig:bad_case_1}). This suggests a failure to effectively parse its own operational history.
    \item \textbf{Failure to Terminate}: A related issue is the inability to recognize overall task completion. Many agents, despite having successfully achieved all essential states, do not issue a stop action. Instead, they continue to perform irrelevant operations until the AITK framework terminates them for exceeding the maximum step limit.
\end{itemize}

This lack of progress awareness points to a deeper deficiency in the agents' planning and state-tracking capabilities. An agent that cannot reliably determine what it has already done or recognize when its final goal has been met will struggle with complex, multi-step tasks. Improving this self-awareness and goal recognition is therefore a crucial direction for future research in developing more robust and efficient GUI agents.

\begin{figure*}
    \centering
    \includegraphics[width=0.9\linewidth]{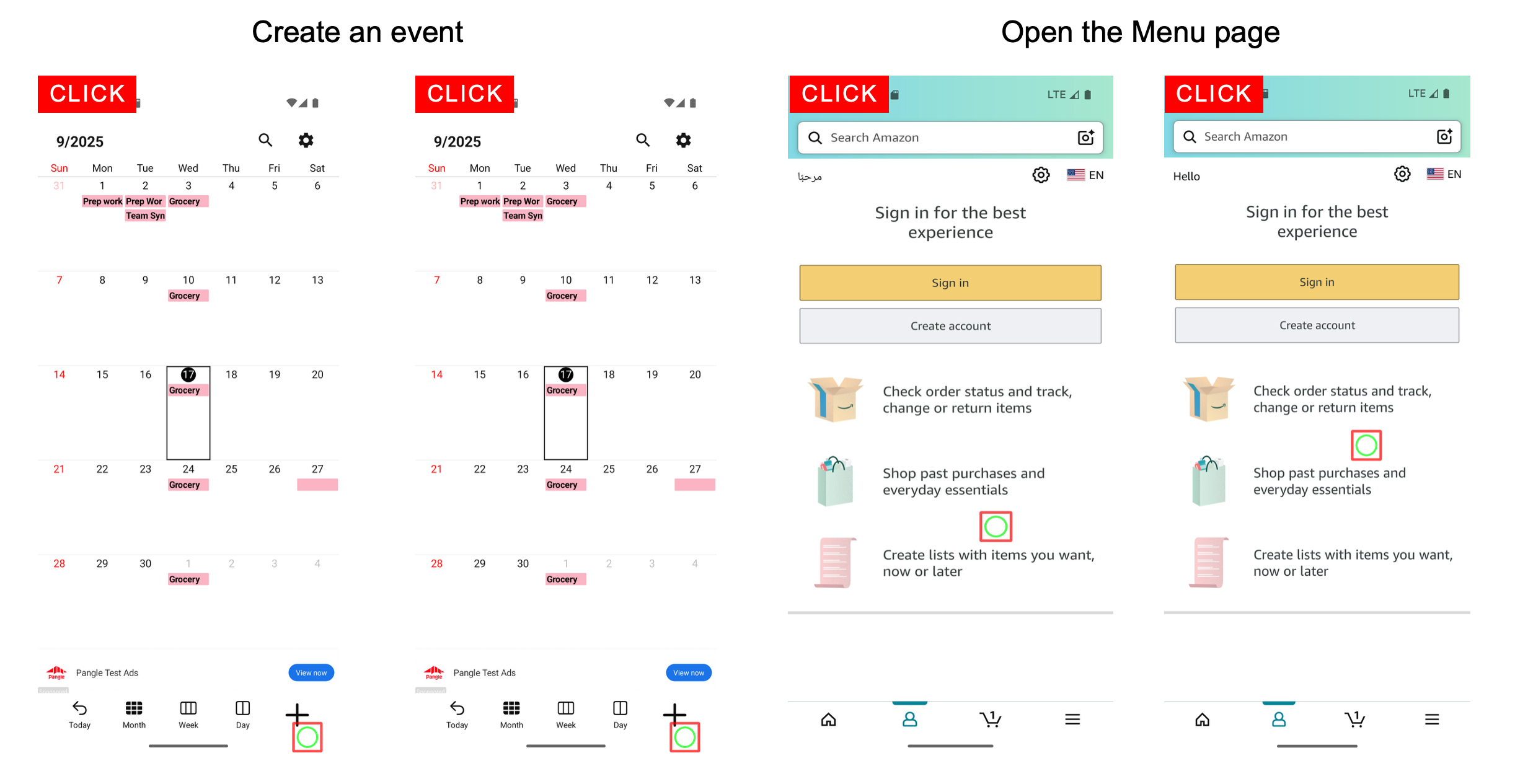}
    \caption{Examples for screen misunderstanding in an essential state.}
    \label{fig:bad_case_3}
\end{figure*}

\paragraph{Screen misunderstanding} Another commonly observed failure mode is a lack of Screen Comprehension, where an agent fails to accurately ground its intended action to the correct visual element on the screen. This issue, which primarily affects \texttt{click} actions, manifests in two distinct ways:

\begin{itemize}[leftmargin=5mm]
    \item \textbf{Incorrect Element Identification:} In some cases, the agent's reasoning is sound (e.g., "click the menu button") but it visually misidentifies the target, instead clicking a different element like a profile icon. This represents a failure in high-level visual recognition, as shown in Figure~\ref{fig:bad_case_3} left.
    \item \textbf{Inaccurate Coordinate Localization:} In other instances, the agent correctly identifies the target element but fails to predict its precise coordinates. This results in a "shifted click" that lands on an adjacent area instead of the intended element, as illustrated in Figure~\ref{fig:bad_case_3} right. This points to a limitation in fine-grained spatial reasoning.
\end{itemize}

\subsubsection{Dynamic Online Apps Specialty}

We analyzed failure cases specific to the dynamic nature of online apps and identified three primary categories of error:

\begin{itemize}[leftmargin=10pt]
    \item \textbf{Information Similarity and Ranking:} Online apps often present search results where the correct item is buried among highly similar distractions. Agents frequently struggle to discern the target from these visual distractors. For instance, in YouTube tasks, agents often select the top-ranked video rather than performing the necessary scrolling actions to locate the specific target content further down the list.

    \item \textbf{Dynamic Interference:} Unlike static offline benchmarks, online environments feature unpredictable elements such as pop-up ads and sponsored content. In apps like Chrome and Home Workout, agents often fail to distinguish between organic results and sponsored links, or lack the logic to identify and close sudden whole-screen ads, leading to task deadlocks.

    \item \textbf{Rich Information and Layout Complexity:} The dense information presentation and deep hierarchical designs of modern apps hinder effective planning. In complex interfaces like Amazon and Omio, agents are often overwhelmed by the volume of product details or struggle to navigate nested filtering and sorting menus, resulting in incorrect navigation paths.
\end{itemize}

\subsection{Generalization Analysis}
\label{app:generalization}

To evaluate the generalization capability of the proposed evaluator and the essential state representation, we conducted experiments on 25 newly introduced tasks outside the original training distribution, comparing Gemini-2.5-pro, A3RM, and A3RM-Continued trained with additional data from the new tasks. As shown in Table~\ref{tab:A3RM_continue}, A3RM still outperforms Gemini-2.5-pro on unseen tasks across all metrics. This result highlights the advantage of a specialized evaluation model trained with structured supervision, demonstrating that A3RM is able to generalize beyond its training distribution to a meaningful extent, benefiting from structured supervision rather than solely relying on the broad reasoning capabilities of general purpose models.

More importantly, the performance gains of A3RM-Continued highlight the extensibility of the essential state representation. With continued training on new tasks, A3RM-Continued achieves consistent improvements in Precision and Accuracy while maintaining stable Recall, indicating that essential states capture transferable structural regularities of GUI task execution. By modeling semantically coherent state transitions and task-relevant state dependencies, essential states provide stable and well-aligned supervision that naturally generalizes to previously unseen task distributions. Consequently, the essential-state framework supports systematic expansion to new tasks, establishing a general and scalable foundation for GUI-Agent evaluation.

\begin{table}[t]
\centering
\resizebox{0.99\linewidth}{!}{
\begin{tabular}{l C{2.8cm} C{1.6cm} C{3.0cm}}
\toprule
\textbf{\quad Metric} & \textbf{Gemini-2.5-pro} & \textbf{A3RM} & \textbf{A3RM-Continued} \\
\midrule
\multicolumn{4}{l}{\textit{Essential State Level}} \\
\quad Precision  & 87.5 & 93.1 & \textbf{97.7} \\
\quad Recall     & 93.3 & 91.1 & \textbf{95.6} \\
\quad F1 Score   & 90.3 & 92.1 & \textbf{96.6} \\
\quad Accuracy   & 87.1 & 90.0 & \textbf{95.7} \\
\midrule
\multicolumn{4}{l}{\textit{Task Level}} \\
\quad Precision  & 93.4 & 94.1 & \textbf{100.0} \\
\quad Recall     & 88.2 & 94.1 & \textbf{94.1} \\
\quad F1 Score   & 90.7 & 94.1 & \textbf{97.0} \\
\quad Accuracy   & 88.0 & 92.0 & \textbf{96.0} \\
\bottomrule
\end{tabular}
}
\caption{Performance comparison on 25 newly introduced tasks.}
\label{tab:A3RM_continue}
\end{table}

\subsection{Risks \& License}

\begin{enumerate}[leftmargin=7mm]
    \item Since the apps are online and some of them requires user login, agent's wrong actions may lead to content malfunctioning. 
    \item MIT license: GUI-OWL, UI-Genie. Apache: UI-TARS, UI-Venus, InfiGUI-R1, Qwen2.5/3-VL. All use are consistent with the license.
\end{enumerate}

\subsection{Training Details}

We use L40s (48G) to train A3RM for 600 GPU-hour. We use EasyR1~\cite{zheng2025easyr1} as the training codebase and use DAPO as the RL training algorithm.

\subsection{AI Declaration}

We use Gemini to help paper writing and one of the evaluation method.

\begin{table*}[t]
    \centering
    \small
    \caption{Sample tasks and their corresponding essential states from the A3 Benchmark.}
    \label{tab:task_examples}
    \renewcommand{\arraystretch}{1.3} 
    
    \begin{tabular}{p{0.45\linewidth} p{0.50\linewidth}}
    \toprule
    \textbf{Task Instruction} & \textbf{Essential States} \\
    \midrule
    
    Navigate to CNN's Science section and check the top headline news. What is the title? & 
    \begin{itemize}[leftmargin=*, nosep, after=\vspace{-\baselineskip}, before=\vspace{-0.5\baselineskip}]
        \item CNN Science section is selected
        \item The top headline news in Science section is selected
        \item The title of the article is answered
    \end{itemize} \\
    \midrule

    Search for 'hotels in Seoul' on Tripadvisor for 1 room and 2 adults, sorted by traveler ranking. & 
    \begin{itemize}[leftmargin=*, nosep, after=\vspace{-\baselineskip}, before=\vspace{-0.5\baselineskip}]
        \item Room is set to 1 room
        \item Guest details are set to 2 adults
        \item 'hotels in Seoul' is searched
        \item Results are sorted by traveler ranking
    \end{itemize} \\
    \midrule

    Open Amazon and search for 'Laptop Sleeve'. Filter the material by 'carbon fiber'. What is the price of the first result? & 
    \begin{itemize}[leftmargin=*, nosep, after=\vspace{-\baselineskip}, before=\vspace{-0.5\baselineskip}]
        \item 'Laptop Sleeve' is searched
        \item Results are filtered by material 'carbon fiber'
        \item Price of the first result is answered
    \end{itemize} \\
    \midrule

    Find the cheapest flight from Paris to Rome for 2 adults on Omio departing tomorrow. What is the total price? & 
    \begin{itemize}[leftmargin=*, nosep, after=\vspace{-\baselineskip}, before=\vspace{-0.5\baselineskip}]
        \item Departure location is set to Paris
        \item Arrival location is set to Rome
        \item Passengers set to 2 adults
        \item Departure date set to tomorrow
        \item Flights are searched
        \item Results for flights are sorted by 'Cheapest price'
        \item Cheapest flight price is answered
    \end{itemize} \\
    \midrule

    Open Citymapper and search route from London Bridge to Oxford Street. How long is the estimated walking time? & 
    \begin{itemize}[leftmargin=*, nosep, after=\vspace{-\baselineskip}, before=\vspace{-0.5\baselineskip}]
        \item Route starting location is set to London Bridge
        \item Route destination is set to Oxford Street
        \item Estimated walking time is answered
    \end{itemize} \\
    \midrule

    Open Gmail and send an email to 'stock\_notify\_01@163.com' with subject 'Meeting time' and body 'When is the meeting?' & 
    \begin{itemize}[leftmargin=*, nosep, after=\vspace{-\baselineskip}, before=\vspace{-0.5\baselineskip}]
        \item New email page is opened
        \item Recipient set to 'stock\_notify\_01@163.com'
        \item Subject 'Meeting time' is added
        \item Body 'When is the meeting?' is added
        \item Email is sent
    \end{itemize} \\
    \midrule

    Search 'How to cook steak' in Youtube, play video by 'Hodder Books' titled 'Gordon Ramsay’s Ultimate Cookery Course' & 
    \begin{itemize}[leftmargin=*, nosep, after=\vspace{-\baselineskip}, before=\vspace{-0.5\baselineskip}]
        \item 'How to cook steak' is searched
        \item Video by 'Hodder Books' with title 'Gordon Ramsay’s Ultimate Cookery Course' is selected
        \item Video is played
    \end{itemize} \\
    \midrule

    Open Home Workout, search Abs workouts, select 'abs beginner', and start training. & 
    \begin{itemize}[leftmargin=*, nosep, after=\vspace{-\baselineskip}, before=\vspace{-0.5\baselineskip}]
        \item Abs workouts are searched
        \item Workout 'abs beginner' is selected
        \item Training is started
    \end{itemize} \\
    \midrule

    Remove butter from shopping list in supercook, mark it as pantry ingredient, and tell me how many recipes can be made. & 
    \begin{itemize}[leftmargin=*, nosep, after=\vspace{-\baselineskip}, before=\vspace{-0.5\baselineskip}]
        \item Shopping list is accessed
        \item Butter is removed from shopping list
        \item Butter is marked as pantry ingredient
        \item Number of recipes possible is answered
    \end{itemize} \\
    \midrule

    Search 'Quantum Computing' in Wikipedia, select first article, open table of contents, and go to Algorithms chapter. & 
    \begin{itemize}[leftmargin=*, nosep, after=\vspace{-\baselineskip}, before=\vspace{-0.5\baselineskip}]
        \item 'Quantum Computing' is searched
        \item First article is selected
        \item Table of contents is opened
        \item Algorithms chapter is accessed
    \end{itemize} \\

    \bottomrule
    \end{tabular}
\end{table*}

\end{document}